%% file: root.tex
\pdfoutput=1

\documentclass[letterpaper, 10pt, conference]{ieeeconf}

\IEEEoverridecommandlockouts
\usepackage{graphicx}
\usepackage{mathtools,physics}

\usepackage{paper}
\usepackage{tabularx}
\usepackage{multirow, multicol}
\newcolumntype{P}[1]{>{\centering\arraybackslash}p{#1}}
\usepackage{booktabs}
\usepackage{multirow}
\usepackage[para]{threeparttable}
\usepackage{dsfont,wrapfig,booktabs}
\usepackage[final]{microtype}
\usepackage{framed}

\usepackage{tcolorbox}
\tcbuselibrary{breakable}

\usepackage{lipsum}
\usepackage{caption}
\usepackage{subcaption}
\usepackage{hyperref}
\hypersetup{
    colorlinks=true,
    linkcolor=blue,
    filecolor=magenta,      
    urlcolor=cyan
}
\usepackage[capitalize]{cleveref}
\crefformat{enumi}{§#2#1#3}
\crefmultiformat{enumi}{§§#2#1#3}{ and~#2#1#3}{, #2#1#3}{, and~#2#1#3}
\crefformat{equation}{(#2#1#3)}
\crefmultiformat{equation}{(#2#1#3)}{ and~(#2#1#3)}{, (#2#1#3)}{, and~(#2#1#3)}
\crefrangeformat{equation}{(#3#1#4)--(#5#2#6)}
\renewcommand{\Cref}[1]{\cref{#1}}

\usepackage{cite}

\usepackage{algpseudocode}
\usepackage{algorithm}

\newcommand\blfootnote[1]{%
  \begingroup
  \renewcommand\thefootnote{}\footnote{#1}%
  \addtocounter{footnote}{-1}%
  \endgroup
}

\title{
An Active Perception Game for Robust Exploration
}
\author{Siming He, Yuezhan Tao, Igor Spasojevic, Vijay Kumar and Pratik Chaudhari
}

\input{notation}
\begin{document}
\fontsize{9.95}{11.9}\selectfont

\input{icra_version}

\input{lemma_appendix}

\bibliographystyle{IEEEtran}
\bibliography{references}

\end{document}

%% file: notation.tex
\def \I {\text{I}}
\def \KL {\text{KL}}
\def \H {\text{H}}

\def \Ym {Y_1^t}
\def \Yp {Y_t^{t+\Delta t}}

\def \ym {y_1^t}
\def \yp {y_t^{t+\Delta t}}

\def \Xp {X_t^{t+\Delta t}}

\def \xp {x_t^{t+\Delta t}}

\DeclareMathOperator*{\maximize}{\text{maximize}}

%% file: icra_version.tex

\clearpage
\maketitle

\begin{abstract}
Active perception approaches select future viewpoints by using some estimate of the information gain. An inaccurate estimate can be detrimental in critical situations, e.g., locating a person in distress. However the true information gained can only be calculated \emph{post hoc}, i.e., after the observation is realized.
We%
\blfootnote{The authors are with the General Robotics, Automation, Sensing and Perception (GRASP) Laboratory, University of Pennsylvania.\\ Email: \{siminghe, yztao, igorspas, kumar, pratikac\}@seas.upenn.edu.}
present an approach to estimate the discrepancy between the estimated information gain (which is the expectation over putative future observations while neglecting correlations among them) and the true information gain. The key idea is to analyze the mathematical relationship between active perception and the estimation error of the information gain in a game-theoretic setting. Using this, we develop an online estimation approach that achieves sub-linear regret (in the number of time-steps) for the estimation of the true information gain and reduces the sub-optimality of active perception systems.
%
We demonstrate our approach\footnote{Code is available at \href{https://github.com/grasp-lyrl/active-perception-game}{https://github.com/grasp-lyrl/active-perception-game}. Proofs are available at \href{https://arxiv.org/abs/2404.00769}{https://arxiv.org/abs/2404.00769}.} for active perception using a comprehensive set of experiments on: (a) different types of environments, including a quadrotor in a photorealistic simulation, real-world robotic data, and real-world experiments with ground robots exploring indoor and outdoor scenes; (b) different types of robotic perception data; and (c) different map representations. On average, our approach reduces information gain estimation errors by 42\%, increases the information gain by 7\%, PSNR by 5\%, and semantic accuracy (measured as the number of objects that are localized correctly) by 6\%. In real-world experiments with a Jackal ground robot, our approach demonstrated complex trajectories to explore occluded regions.
\end{abstract}

\section{Introduction}

\begin{figure}
\centering
\includegraphics[width=\linewidth]{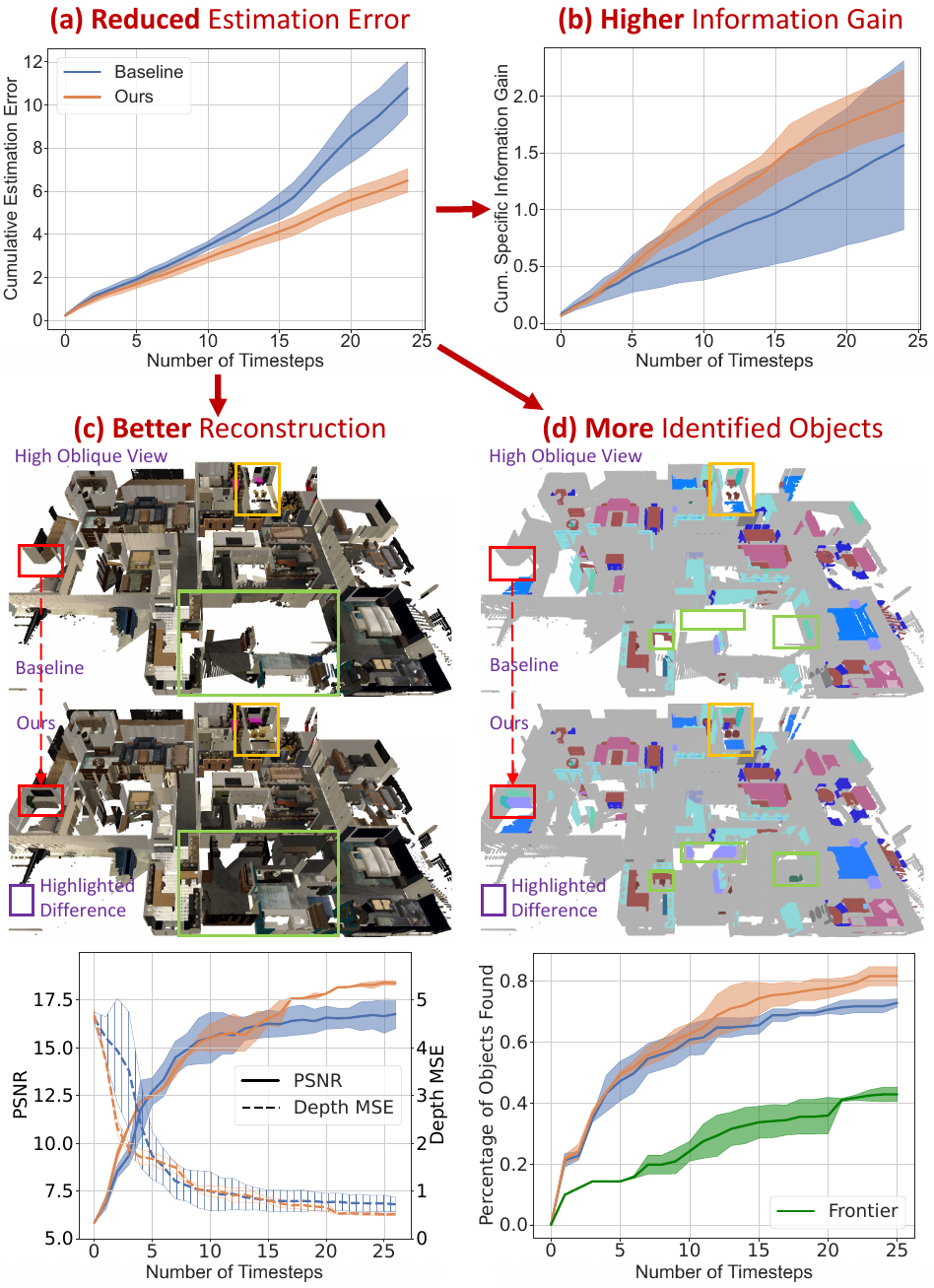}
\caption{
\textbf{Comparison of our approach for active perception against a baseline for a quadrotor exploring an indoor environment in a photorealistic simulator.} Our approach reduces estimation errors (a), leads to a higher information gain (b), better reconstruction with higher peak signal-to-noise-ratio (PSNR) and lower depth mean square error (MSE) in the learned neural radiance field (NeRF) (c), and leads to an increase the total number of objects that are correctly localized in the scene (d).
}
\label{fig:firstfigure}
\end{figure}

As robots transition from controlled laboratory environments to real-world settings, their ability to actively perceive information from their surroundings becomes increasingly crucial.
Active perception has gained research interest due to its potential applications in search and rescue, planetary exploration, environmental monitoring, and structural inspection \cite{placed2023survey, viewplanning2020}.

An active perceiver knows why it wishes to perceive, and chooses what, how, when and where to achieve that perception~\cite{5968, bajcsy2016revisitingactiveperception}. We instantiate active perception as taking actions to maximize information gain. The notion of information gain can vary depending on the application. For example, in search and rescue, the robot may need to acquire the geometric information about a collapsing structure or collect photometric and semantic information to identify people in distress and assess their health conditions.
Regardless of the specific definition, the ``true'' information gain is the incremental knowledge obtained from future observations.
Since future observations are unknown \emph{a priori}, the robot must rely on estimates to guide its decisions. However, correlations in observations, model uncertainty, and data uncertainty can lead to inaccurate estimates, resulting in suboptimal viewpoints. This is especially critical in safety-sensitive tasks like search and rescue, where missing vital information can have severe consequences.

To address the issue, this paper proposes an online procedure for estimating the actual information gain. The key mathematical idea is as follows. We model active perception as a game where the robot selects informative viewpoints while the adversarial player provides inaccurate estimates of the information gain at these viewpoints to mislead the robot. Since the environment is fixed, the adversarial player cannot cause arbitrarily large harm. We design an online optimization algorithm that allows the robot to learn from the discrepancy between its pre-observation estimates and the actual information gained after receiving the observations. This learning process enables the robot to fix its inaccurate estimates and improve active perception performance in the subsequent time-steps as shown in~\cref{fig:firstfigure}. This mathematical approach provides explicit guarantee on the regret of both (a) the quality of the estimated information gain $O(T^{3/4})$ in the length of the episode $T$, and (b) the quality of the entire active perception pipeline including informative path planning $O(T^{3/4} + \lambda T + \Delta)$ where $\lambda$ and $\Delta$ is described in~\cref{sec:algorithm}.

Our approach is very general. It can be used to improve the performance of any active perception procedure. We demonstrate our approach for active perception using a comprehensive set of experiments on: (a) different types of environments, including a quadrotor in a photorealistic simulation, real-world robotic data, and real-world experiments using ground robots exploring indoor and outdoor scenes; (b) different types of robotic perception data; and (c) different map representations.

\section{Related Work}
\underline{Methods with performance guarantees:} Active sensing in Gaussian process (GP) has been well studied in \cite{10.5555/1390681.1390689}. Finding optimal sensor placement has intractable runtime. The work obtains near-optimal online bounds for greedy algorithms based on the submodularity of mutual information. The analysis assumes GP has a known covariance function which is usually unrealistic in real-world applications. An extension \cite{krause2007nonmyopic} uses a near-optimal exploration-exploitation approach for stationary GP with an unknown covariance function. For non-stationary GP, a single estimation of covariance function is not enough. The paper \cite{krause2007nonmyopic} proposed manually dividing the whole space into smaller areas with a stationary process. Then, a similar analysis also gives an efficient and near-optimal algorithm for informative path planning in GP \cite{10.5555/1622716.1622735,6224902,s19051016}.

\underline{Methods without performance guarantees:} Occupancy map is commonly used in robotics tasks, but active perception algorithms based on occupancy maps usually don't provide performance guarantee analysis. Next-best-views (NBV) methods samples and selects viewpoint that maximizes some utility functions. Utilities based on the amount of observed volume is used in \cite{1087372,7487281,Yoder2016}. The following works \cite{Border_2018, Schmid_2020} used surface reconstruction utility in addition to volumetric utility. Information-theoretic methods \cite{8793541,7139865,9196592} design efficient algorithms to calculate mutual information between range sensor data and occupancy maps. Since the occupancy of unseen areas is unknown, NBV and information-theoretic methods could underestimate or overestimate the utility. To address this issue, several works \cite{8793769,tao2023learning,Tao_2023} propose map completion. The methods train neural networks to predict the occupancy of unseen areas based on existing datasets. Semantic and learned scene representations become popular in recent years and have been used for loop-closure, navigation, and decision making. Uncertainty in those representations is also used as a utility for exploration \cite{georgakis2022learning,he2023active, ctx55653014670003681, hsu2024active, asgharivaskasi2023semantic, tao2024rt}.
Reinforcement learning methods \cite{Tao_2023, dosovitskiy2017learning,sun2022multiagent, chaplot2020learning} are used by learning to actively perceive based on past data. Information gain estimations in occupancy, semantic, and learned maps could be inaccurate due to uncertainties described in~\cref{sec:algorithm}, and the performance guarantee would not hold. This gap motivates our design of accurate information gain estimator and analysis of performance guarantee for general scene representations.

\section{Method}

\subsection{Problem Formulation}
\label{sec:formulation}
Let $\Theta$ denote a fixed scene from which a robot obtains a sequence of measurements $Y_1^t = (Y_1, \dots, Y_{t-1})$ from viewpoints $X_1^t = (X_1, \dots, X_{t-1})$ respectively where each $X_k \in \text{SE}(3)$ and each measurement $Y_k$ is an RGBD image.
We can write $p(Y_t \mid X_t, \Theta)$ as the probability of obtaining an observation $Y_t$ from viewpoint $X_t$. At time $t$, the robot selects the next $\Delta t$ viewpoints $X_t^{t+\Delta t}$ to obtain measurements $Y_t^{t+\Delta t}$.
Access to measurements gives us information about the scene via the conditional probability $p(\Theta \mid \Ym)$.\footnote{We will use capital letters $X, Y$ to denote random variables and small letters $x, y$ to denote their corresponding values.}
The discrepancy between the intrinsic uncertainty of the scene $p(\Theta)$ and this conditional probability (averaged over past measurements) is given by the mutual information
\begin{equation}
    \I(\Theta; \Ym) = \int \dd{p(\ym)} \KL(p(\Theta \mid \ym), p(\Theta))
    \label{eq:mi}
\end{equation}
where $\KL(\cdot, \cdot)$ denotes the Kullback-Leibler divergence, and $\I$ denotes mutual information. The expected information gain, i.e., the incremental amount of information obtained from new measurements (given past ones $\ym$) is
\begin{equation}
    \begin{aligned}
        r_t &\triangleq \I(\Theta; \Yp \mid \ym)\\
        &= H(\Theta \mid \ym) - H(\Theta \mid \Yp, \ym).
    \end{aligned}
    \label{eq:ig}
\end{equation}
where $H$ denotes the Shannon entropy.%
\footnote{Note that $H(\Theta \mid \ym) = -\int \dd{p(\theta \mid \ym)} \log p(\theta \mid \ym)$ while $H(\Theta \mid \Yp, \ym)$ equals
\[
    -\int \dd{p(\yp \mid \ym)} \dd{p(\theta \mid \yp, \ym)} \log p(\theta \mid \yp, \ym).
\]
}
It is important to emphasize that expected information gain is a function of the past measurements that were realized $\ym$, not the random variable $\Ym$. It is also a function of the distribution of the future viewpoints $\Xp$ chosen by the robot. To emphasize:
\[
    r_t \equiv r_t(\Xp \mid \ym).
\]
At each time-step, the robot selects future viewpoints using past measurements $\Xp \mid \ym$ to maximize the information gain. It solves the optimization problem
\begin{equation}
    \maximize_{\Xp \mid \ym}\; r_t
    \label{eq:problem}
\end{equation}
by estimating the information gain.

\begin{framed}
\noindent \textbf{The key idea of this paper is as follows.} After visiting future locations $\xp$, the robot obtains new measurements $\yp$. It can now calculate the specific information gain, i.e., the realized reduction of uncertainty about the scene,
\begin{equation}
    r^*_t \triangleq H(\Theta \mid \ym) - H(\Theta \mid \yp, \ym);
    \label{eq:rstar}
\end{equation}
note the difference with respect to~\cref{eq:ig} where the conditioning is on $\Yp$.
The discrepancy 
\begin{equation}
    \begin{aligned}
    \Delta r_t &= r_t - r^*_t\\
    &= H(\Theta \mid \Yp, \ym) - H(\Theta \mid \yp, \ym).
    \end{aligned}
    \label{eq:dig}
\end{equation}
will lead to a suboptimal choice of viewpoints.
In addition to the expected information gain $r_t$, the robot can estimate---in hindsight---the discrepancy
and thereby hope to maximize a quantity closer to the specific information gain in the next step.
\end{framed}

\subsection{Estimating the specific information gain}
\label{sec:algorithm}
The robot's estimate of the specific information gain $r^*_t$, given the position $s$ in the ordered sequence of $\Delta t$ observations and the expected information gain $r$, is denoted by
\begin{equation}
    \hat r^* = f(s, r),\text{ where }
    f: \mathbb{N} \times \mathbb{R} \mapsto \mathbb{R}.
    \label{eq:hat_r}
\end{equation}
This function $f$, which we call the \textbf{improvement function}, is fitted using past expected information gains $r_1^t$ and past discrepancies $\Delta r^t_1$. 
%
The improvement function can mitigate the discrepancy between expected information gain and specific information gain:
\begin{enumerate}
\item[(a)] \underline{Correlations in successive observations}: If the robot chooses the next $\Delta t$ viewpoints independently, its estimate of $\I(\Theta; \Yp\mid \ym)=\sum^{\Delta t-1}_{i=0} \I(\Theta; Y_{t+i} \mid y^{t+i-1}_1)$ in~\cref{eq:ig} will be computed using $\sum^{\Delta t-1}_{i=0} \I(\Theta; Y_{t+i} \mid \ym)$. The two are not identical when observations are correlated with each other, e.g., when views overlap. In general, the latter is an overestimation. The first argument $s$ of our improvement function adjusts for discrepancies $\I(\Theta; Y_{t+i} \mid y^{t+i-1}_1) - \I(\Theta; Y_{t+i} \mid \ym)$ due to correlations.
\item[(b)] \underline{Model (epistemic) uncertainty}: The scene representation $\Theta$ is learned from past observations. If the representation is insufficient to accurately capture the true scene, for example, due to limited data or inadequate training, discrepancies arise between $\H(\Theta \mid \Yp, \ym )$ and $\H(\Theta \mid \yp, \ym )$. The second argument $r$ adjusts for these discrepancies by identifying over- and underestimates.
\item[(c)] \underline{Data (aleatoric) uncertainty}: For example, in a dynamic scene, say a propeller with spinning blades, each voxel's occupancy depends upon the angular velocity of the blade. This results in consistently high expected information gain estimation but low specific information gain, as the robot realizes after the observation that the uncertainty is not reducible. A robot that maximizes the expected information gain will never look away from the scene even if there is not much information to gleaned from it. The second argument $r$ identifies and accounts for the overestimates of reducible uncertainty $r^*_t$.
\end{enumerate}

\textbf{Representing the improvement function:}
We query the improvement function $f(s,r)$ in~\cref{eq:hat_r} to estimate specific information gain of future viewpoints corresponding to a horizon of length $\Delta t$, so the first argument has domain $s \in \{0,\dots,\Delta t-1\}$. Let us suppose that the expected information gain is upper bounded by $\beta$, and therefore the domain of the second argument is $[0, \beta] \subset \mathbb{R}$ which we discretize into $b$ equal-sized bins $[(k-1) \beta/b, k \beta/b)$ for $k \in \{1,\dots,b\}$. We can now represent the function $f$ as a matrix
\begin{equation}
    f \in \mathbb{R}^{\Delta t \times b}.
    \label{eq:f}
\end{equation}
This matrix is updated by the robot after every $\Delta t$ observations using the specific information gain as follows. The robot maintains a variable
\[
    \alpha_{s,i} = \sum_{s'=1}^t \mathds{1}\{r_{s'} \in \text{ bin } i\}\quad \forall s \in [t, t+\Delta t-1].
\]
The improvement function is updated for all $s \in [t, t+\Delta t-1]$ and all $i \in \{1,\dots,k\}$ as
\begin{equation}
    \begin{aligned}
    f_{s,i} \leftarrow
            r_{s, i} +
            \sqrt{\frac{\alpha_{s-\Delta t,i}}{\alpha_{s,i}}} (f_{s,i} -r_{s,i})
            - \frac{\beta \text{sign}(f_{s,i} - r^*_{s,i})}{4 \sqrt{\alpha_{s,i}}}
    \end{aligned}
    \label{eq:fupdate}
\end{equation}
where $r_{s, i}$ highlights that the information gain at timestep $s$ is in the bin $i$.
The next section proves that this update rule for the improvement function achieves sublinear regret $O(T^{3/4})$ for estimating the specific information gain. The first term is the most elementary estimate of the specific information gain; it is simply $r_{s,i}$. The second term downweights the influence of prior learning, allowing the third term to emphasize the current update. The third term fixes the discrepancy between the estimated specific information gain and the actual specific information gain in hindsight, with a weight that is proportional to the learning rate to the element $f_{s,i}$ of the matrix $f$.

\textbf{Selecting viewpoints using the estimated specific information gain}
Viewpoints $X_t$ should be selected using~\cref{eq:problem} where $r_t$ is replaced by the estimation of specific information gain in~\cref{eq:fupdate}. But, even if we have a more accurate estimate of the information gain now, the optimization problem in~\cref{eq:problem} (also called the ``informative path planning problem'') is computationally intractable.
There is existing work that has studied such problems to obtain near-optimal algorithms~\cite{10.5555/1390681.1390689}. For instance, the eSIP algorithm in~\cite{10.5555/1622716.1622735} selects viewpoints $X_t$ that satisfy
\[
    \sum_{t=1}^T r^*_t(X_t \mid \ym) \geq \frac{(1-1/e)}{(1+\log_2 \Delta t)} \sum_{t=1}^T r^*_t(X^*_t\mid \ym),
\]
where $X_t^*$ is the optimal solution of the optimization problem and $e$ is Euler's number 2.718. We will see in the next section that if our viewpoint selection also satisfies a similar inequality:
\[
    \sum_{t=1}^T r^*_t(X_t\mid \ym) \geq \gamma \sum_{t=1}^T r^*_t(X^*_t\mid \ym)
\]
for some $\gamma \in (0,1]$, then we can bound the sub-optimality of the information gain of our approach (which estimates the specific information gain $r_t^*$ using the expected information gain $r_t$) to be $O(T^{3/4} + \lambda T + \Delta)$ in~\cref{thm:4.4}. Here $\lambda$ bounds the maximum discrepancy between the specific information gain in each $(s,i)$ and the optimal estimation achievable within the constraints of the function class. The additive term  $\Delta$ is the regret coming from informative path planning.

\section{Analysis}
\label{sec:analysis}
We model the discrepancy $\Delta r_t$ as an adversary, meaning that it can significantly mislead active perception algorithms if decisions rely solely on $r_t$. Treating the discrepancy as adversarial allows the analysis to hold under minimal assumptions. We frame active perception as a game between the robot and this adversary: the robot aims to navigate to the most informative viewpoints, while the discrepancy attempts to mislead it. Our algorithm's analysis is grounded in this game-theoretic framework.

\subsection{Bound on regret of online estimation with full feedback}\label{section:predeictionregret}
We first analyze our algorithm in a full-information setting, where the specific information gain is received for all possible paths. We will then demonstrate a  reduction from the actual setting where specific information gain is received only for the path executed by the robot, this is the so-called bandit feedback.

Each element of the matrix $f$ is learned independently, so we focus the analysis on a single element $f_{s,i}$. We skip time-steps that do not update $f_{s,j}$ and denote the others using $a \in \{1,\cdots, \alpha^*_{s,i}\}$ where $\alpha^*_{s,i}$ is the total number of updates to $f(s,i)$. For brevity, we will use shorthand $f_a$ (improved estimate), $r_a$ (elementary estimate), $r^*_a$ (specific gain), $\alpha_a$ (number of updates), and $d_a$ (sign of the error $f_{s,i} - r^*_{s,i}$)~\cref{eq:fupdate}. These quantities change with time.

Let $\delta_a \equiv f_a - r_a$. Online prediction in the full-information setting is modeled as an online convex optimization problem. For each timestep $a$, the robot selects $\delta_{a}$ to minimize the loss $\ell_a = \vert \Delta r_{a} - \delta_{a}\vert$. When the robot finishes active perception and looks back, it realizes $\delta^*$ is the estimation that it should have made. Regret is the additional loss incurred by estimating $\delta_a$ instead of $\delta^*$:
\begin{equation}
    \rho_{s,i} = \sum^{\alpha^*_{s,i}}_{a=1} \ell_a  - \ell^*_{s, i} \text{ where }
    \ell^*_{s, i} = \min_{\delta^*}\sum^{\alpha^*_{s,i}}_{a=1}\left\vert \Delta r_{a} - \delta^* \right\vert
    \label{eq:subgroupregret}
\end{equation}
The overall regret is
\begin{equation}
    \rho = \sum^{\Delta t}_{s=1}\sum^b_{i=1} \rho_{s,i}. 
    \label{eq:overallbound}
\end{equation}
The update rule in~\cref{eq:fupdate} is a special case of ``follow the regularized leader'', see~\cref{lemma:4.1} in the Appendix. This can be used to bound the regret in~\cref{lemma:4.2} in the Appendix, which leads to the following theorem.
\begin{theorem}\label{thm:4.1}
    For the ``follow the regularized leader'' learning rate $\eta_{a} = \beta/\sqrt{\alpha_a}$ in~\cref{eq:fupdate}, the regret is
    \[
    \rho_{s,i} \leq \beta(N\sqrt{\alpha^*_{s,i}}+1)\leq \beta(N\sqrt{T}+1),
    \]
    where the robot selects from among $N$ candidate viewpoints at each timestep. The bound on $\rho$ is in worst case $\Delta t b \beta (N\sqrt{T}+1)$.
\end{theorem}
To prove, plug in the specified $\eta_{a}$ to the bound in~\cref{lemma:4.2}. The regret $\rho_{s,i}$ is bounded due to Holder’s inequality $\sum^{\Delta t}_{s=1}\sqrt{1/s} \leq 2\sqrt{\Delta t}$.


\subsection{Bound on regret of online estimation with bandit feedback}\label{subsec:banditfeedbackregret}
Using the approach of~\cite{Slivkins}, bandit-feedback setting can be reduced to the full-information setting.
For $a=1,\dots,\alpha^*_{s,i}$, let the probability that $x_a$ is on the executed path be $p_{a}$. The hallucinated loss $\hat \ell_a$ is
\begin{equation}
    \hat \ell_a = \frac{\vert \Delta r_a - \delta_a \vert}{p_a} \mathds{1}(\text{$x_{a}$ is executed})
\end{equation}
This loss is an unbiased estimator of the actual loss, as $\mathbb{E}(\hat \ell_a) = p_a |\Delta r_a - \delta_a|/p_a + 0 = |\Delta r_{a} - \delta_a|$. Additionally, to ensure that this loss is well-defined, we require $p_a$ to be strictly positive. We can achieve this by randomly sampling paths with probability $\tau \in (0, \frac{1}{2})$ in the active perception algorithm.%
\footnote{These randomly sampled paths are necessary for the randomization-based reduction. In practice, the robot does not implement these random paths.}
Since a feasible $y_a$ must be included in at least one possible path, $y_{a}$ is selected with a probability of at least $N^{-\Delta t}$. Given that the algorithm randomly samples paths with probability $\tau$, $p_a$ is greater than or equal to $\tau N^{-\Delta t} \geq 0$. Intuitively, in the bandit-feedback setting, the robot explores the specific information gain of different measurements via randomization. After randomization, we prove the expected regret in bandit-feedback setting in~\cref{lemma:banditregret} in the Appendix. With randomization, the algorithm has sub-linear estimation regret:
\begin{theorem}\label{thm:boundbanditregret}
For $\tau=T^{-1/4}$, the regret bound is
\begin{equation}
    \mathbb{E}(\rho_{s,i}) \leq N^{\Delta t} \beta(NT^{3/4} + T^{1/4}) + \beta T^{3/4}=O(T^{3/4}).
\end{equation}
The overall regret is also $O(T^{3/4})$.
\end{theorem}
To prove, plug in the specified $\tau$ into the bound in~\cref{lemma:banditregret}.

\subsection{Regret of active perception}\label{subsec:apgregret}
This section characterizes the relationship between errors in estimation of the specific information gain, optimality of path planning, and the performance of active perception. We define the additional information gain that could be obtained if the robot plans optimal paths based on specific information gain $r^*_t$ as the active perception regret:
\begin{equation}
    \varrho=\sum_{t=1}^T \left[r^*_t\left(\hat{X}^{t+\Delta t}_t\right) - r^*_t\left(\Xp\right)\right]
\end{equation}
where the viewpoints $\hat{X}^{t+\Delta t}_t$ are selected by maximizing specific information gain and $\Xp$ are selected by maximizing the estimates of specific information gain.
We first show that our estimations of specific information gain lead to bounded active perception regret.
\begin{theorem}\label{thm:4.3}
Given estimation regret bound $\rho$, the bound on active perception regret is given by
\begin{equation}
    \mathbb{E}(\varrho)\leq 4(\rho + \rho')
\end{equation}
where $\rho'=\sum^{\Delta t}_{s=1}\sum^b_{i=1} \ell^*_{s, i}$.
\end{theorem}

Then, we can take the optimality of path planning algorithms into account.
\begin{theorem}\label{thm:4.4}
Given an informative path
planning algorithm which selects $\Xp$ with $\sum_{t=1}^T r^*_t(X_t\mid \ym) \geq \gamma \sum_{t=1}^T r^*_t(X^*_t\mid \ym)$, the active perception regret is
\begin{equation}
    \mathbb{E}(\varrho) \leq 4(\rho + \rho') + \Delta
\end{equation}
where $\Delta=(1-\gamma)\sum_{t=1}^T r^*_t(\hat{X}^{t+\Delta t}_t)$.
\end{theorem}
\begin{remark}\label{rmk:4.1}
With optimal path planning, active perception regret bound is $O(T^{3/4}+\lambda T)$. And with approximate path planning, the regret bound is $O(T^{3/4}+\lambda T+\Delta)$. The bound $\rho$ is shown in~\cref{thm:boundbanditregret} to be sub-linear. If each $\left\vert \Delta r_{a} - \delta^* \right\vert$ in~\cref{eq:subgroupregret} is bounded by $\lambda$, then we have $\rho' \leq nbT\lambda$.
\end{remark}

\section{Experimental Validation}
\label{sec:3}

To evaluate the effectiveness and generalizability of our approach, we conducted experiments on two active perception systems: (i) a semantic neural radiance field (NeRF)-based active perception system on a quadrotor in a photorealistic simulator described in~\cref{subsec:nerf}, and (b) an occupancy map-based active perception system on a ground robot. We tested the effectiveness of our approach for online estimation of specific information gain using a real-world dataset (described in~\cref{subsec:dataexp}). Finally, we demonstrate real-world experiments with the ground robot with all our algorithms running fully onboard in~\cref{subsec:occmap}.


\subsection{Simulation experiments}
\label{subsec:nerf}

\begin{figure}
    \centering
    \includegraphics[width=\linewidth]{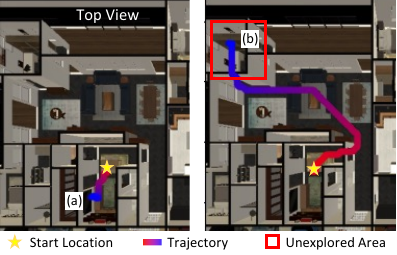}
    \caption{\textbf{Viewpoint selection comparison.}
    Given a set of candidate trajectories, our method selects trajectory (b), which leads to an underexplored room, while the baseline selects trajectory (a), which remains within the same room.
    }
    \label{fig:explored_area}
\end{figure}

\begin{table*}[!t]
\centering
\renewcommand{\arraystretch}{1.25}
\resizebox{\linewidth}{!}{
\scriptsize
\begin{threeparttable}
\begin{tabular}{lrrrrrrrrr}
 & \multicolumn{3}{c}{\textbf{Simulation Scene 2}} & \multicolumn{3}{c}{\textbf{Simulation Scene 3}} & \multicolumn{3}{c}{\textbf{Simulation Scene 4}} \\
\toprule
& Frontier & Baseline & Ours & Frontier & Baseline & Ours & Frontier & Baseline & Ours \\
\textbf{Error\tnote{1}} &   -  &  3.90 $\pm$ 0.55  &    2.95 $\pm$ 0.28 &   -  & 7.85 $\pm$ 4.52  &    3.88 $\pm$ 1.63     &   -  & 3.22 $\pm$ 0.39 &  2.44 $\pm$ 0.18    \\
\textbf{Info. Gain\tnote{2}} &  -  & 1.87 $\pm$ 0.18 &   1.90 $\pm$ 0.23 &   - & 1.25 $\pm$ 0.15  &    1.29 $\pm $ 0.22    &   - & 1.58 $\pm$ 0.15 &     1.54 $ \pm $ 0.01         \\
\textbf{PSNR $\uparrow$} &  -  & 19.53 $\pm$ 0.65 &  20.73 $\pm$ 0.46 &   - & 16.68 $\pm$ 0.89  &    17.12 $\pm $ 0.65    &   - & 15.11 $\pm$ 0.22 &     15.29 $ \pm $ 0.26         \\
\textbf{Depth MSE} &  -  & 0.62 $\pm$ 0.07 &   0.54 $\pm$ 0.07 &   - & 0.89 $\pm$ 0.24  &    0.85 $\pm $ 0.25    &   - & 1.09 $\pm$ 0.09 &     1.13 $ \pm $ 0.09        \\
\textbf{\# Objects (\%)\tnote{3}}  &  62.60 $\pm$ 3.00 & 78.00 $\pm$ 7.20 &   86.20 $\pm$ 2.30 &   55.90 $\pm$ 12.0  & 77.50 $\pm$ 2.00  &    77.50 $\pm$ 1.10     &   62.00 $\pm$ 9.00 & 88.40 $\pm$ 3.80  &   89.90 $\pm$ 2.90\\
\bottomrule
\end{tabular}
\begin{tablenotes}
   \item [1] The mean absolute error of specific information gain estimation. 
   \item [2] The mean of specific information gain.
   \item [3] The percentage of objects successfully located.
 \end{tablenotes}
  \end{threeparttable}
}
\caption{\textbf{Quantitative performance comparison across additional simulation scenes.} 
These results complement~\cref{fig:firstfigure}, further demonstrating the effectiveness of our method in terms of estimation error, information gain, reconstruction, and object identification, as detailed in~\cref{subsec:nerf}.
}
\label{tab:benchmark}
\end{table*}

\subsubsection{Setup}
Neural radiance fields (NeRFs) build a map of the environment capable of rendering images from new viewpoints~\cite{li2023nerfacc}. We build upon the work of~\cite{he2023active} which built a semantic NeRF that can also predict semantic segmentation maps. We do not go into the details of the architecture for lack of space, the reader is encouraged to read these two original papers.%
\footnote{Observations $\yp$ include RGB images, depth measurements and semantic segmentations. We model the color and depth of each ray as a Gaussian distribution estimated from the ensemble. For calculating quantities like $H(\Theta \mid \ym)$, we use estimates of the entropy for these Gaussians. For instantiating our approach with its theoretical guarantees, these quantities should be bounded. We therefore clip the entropy (to a maximum value of 5). Semantic segmentation generates a distribution over categories and its entropy is already bounded. The transmittance of a ray is modeled as a Bernoulli random variable indicating whether the ray hits an obstacle, with entropy between 0 and 1.}
Our goal in this experiment will be to demonstrate the improvements in the quality of the trained semantic neural radiance field (NeRF) using our approach for active perception.

We use a photorealistic 3D simulator called Habitat-Sim~\cite{habitat19iccv, szot2021habitat, puig2023habitat}. The robot starts at a random location in the map and collects an initial set of measurements to initialize the map. The exploration phase begins after this initialization. At each time-step, we sample 20 different locations on the 2D plane, generate paths to these locations using Dijkstra's algorithm, and discretize each path  into $\Delta t$ steps to calculate the estimated specific information gain. The path with the highest information gain estimates is used to generate a dynamically feasible trajectories for the quadrotor using Rotorpy~\cite{folk2023rotorpy}.

An ensemble of NeRFs is used to represent the scene $\Theta$. We calculate the information gain independently for the different observation modalities (RGB, depth, semantic segmentation, and transmittance along a ray which is calculated from depth) and sum them up to estimate the total specific information gain. The number of bins $b$ is set to 100, and the number of measurements $\Delta t$ is set to 40. After execution the path to the chosen viewpoints, the robot computes specific information gain and updates the NeRF in an online fashion using the realized observations.



\subsubsection{Baselines and Metrics}
We conducted experiments with frontier based exploration~\cite{613851}, the approach of~\cite{he2023active} which maximizes the expected information gain and our proposed approach (which estimates the specific information gain) on four simulated environments of varying sizes and complexities. The results are shown in~\cref{fig:firstfigure} and~\cref{tab:benchmark}. Each experiment is repeated three times with the same initialization to obtain the mean and standard deviation of all performance measures. For each experiment, we recorded the mean absolute error of estimation of specific information gain, the mean of specific information gain, peak signal-to-noise ratio (PSNR) and depth mean square error (MSE) of NeRF, and the percentage of objects found during the experiment was recorded.%
\footnote{We do not use coverage because our approach maximizes photometric, geometric and semantic information recovered from the scene and not the area covered.}

\subsubsection{Results}
The proposed approach reduced the error of estimation by 24.2\% - 50.6\%, gathered 0\% - 25.6\% more information, and located 0\% - 12\% more objects compared to the baseline approaches. Compared to frontier-based exploration with a random frontier, our approach located 37.7\% to 100\% more objects. For NeRF reconstructions, our approach achieved 1.2\% to 9.7\% higher PSNR and -3.7\% to 29.2\% lower depth MSE across four scenes. Additionally, our approach has smaller standard deviations for most metrics, which demonstrates that it is consistently better. The advantage of our approach is clearer in Scene 1 and Scene 2 which are larger and more complex. We believe that approaches like ours for estimating the specific information gain are more effective for environments that are large in scale and complexity.

Qualitatively,~\cref{fig:firstfigure}(c) compares explored areas by different approaches. Several rooms that were missed by the baseline are explored by our approach.~\cref{fig:firstfigure}(d) compares the objects located by the baseline and by our approach. Objects of different categories are associated with different colors. Our approach identifies more objects, including small chairs under the table and a toilet in a small bathroom. In~\cref{fig:explored_area}, we give an example of viewpoint selection by baseline and our approach. In one of the timestep, the robot samples a set of candidate trajectories and select the trajectory with highest information gain. By better estimating the specific information gain, our approach selects a trajectory to an underexplored room.

\subsection{M3ED Dataset}
\label{subsec:dataexp}

\begin{table}[!t]
\centering
\renewcommand{\arraystretch}{1.25}
\resizebox{\linewidth}{!}{
\scriptsize
\begin{threeparttable}
\begin{tabular}{lrrrrrr}
& \multicolumn{3}{c}{\textbf{Indoor: building\_loop}}               & \multicolumn{3}{c}{\textbf{Outdoor: penno\_short\_loop}}\\
\toprule
Noise Type $\to$              & None  & Gaussian & Impulse & None & Gaussian & Impulse \\
\textbf{Baseline}          &     0.475     &       0.484         &        0.490       &     0.474     &       0.484         &       0.488        \\
\textbf{Ours}          &     0.163     &        0.160        &        0.162       &     0.162     &       0.163       &       0.172        \\ \bottomrule
\end{tabular}
\end{threeparttable}
}
\caption{\textbf{Estimation improvement for M3ED dataset.} We compare the mean estimation errors of the baseline and our methods under different types of depth measurement noise described in~\cref{subsec:dataexp}.}
\label{tab:dataset}
\end{table}

M3ED~\cite{Chaney_2023_CVPR} is a high-quality dataset that consists of a large set of natural scenes including cars driving in urban, forest, day and night conditions. Since this data is already collected, we cannot perform ``exploration'' in it. But we can use it to evaluate our approach for estimating the specific information gain. This is important because it shows that we can demonstrate our approach on real-world data.

\subsubsection{Setup}
We use an 3D occupancy-map-based map in this section (instead of a NeRF like the previous section). We treat LiDAR data as the ground-truth depth, and LiDAR-based odometry in M3ED as the ground-truth location of the robot. All voxels are initially assigned a probability of occupancy of 0.5. For each measurement and its associated odometry, ray-casting was performed, followed by a log-odds-based map update. To calculate the information gain, rays were uniformly sampled from each viewpoint. Given depth measurements $y_t$ and the sensor model, we calculated the information gain as $H(\Theta) - H(\Theta \mid y_t)$, where $p(\Theta)$ and $p(\Theta \mid y_t)$ represent the probability of occupancy before, and after, map update. Information gain is the sum of information gains from all sampled rays at the given viewpoint. After each realized observation, we calculated the specific information gain and the mean estimation error in the map occupancy. For this experiment, the upper bound on entropy $\beta = 1$, the number of bins $b = 100$, and $\Delta t = 1$.

\subsubsection{Results}
Our approach reduces the error of estimating the specific information gain estimations by over 66\% in both indoor and outdoor scenes in M3ED dataset. See~\cref{tab:dataset}.%
\footnote{Unmodeled noise can introduce aleatoric uncertainty, which is often underestimated in information gain predictions. We added Gaussian noise (zero mean, standard deviation equal to 1/8$^{\text{th}}$ of the raw depth) and Impulse noise (half of the LiDAR observations are replaced with random values between $1/3$ and $5/3$ of the raw depth) to simulate unmodeled sensor noise in LiDAR data. \cref{tab:dataset} shows that the proposed approach significantly improves the estimate of the information gain for unmodeled sensor noise.
}

\subsection{Real-world experiments on ground robots}
\label{subsec:occmap}

\begin{figure}
     \centering
     \includegraphics[width=\linewidth]{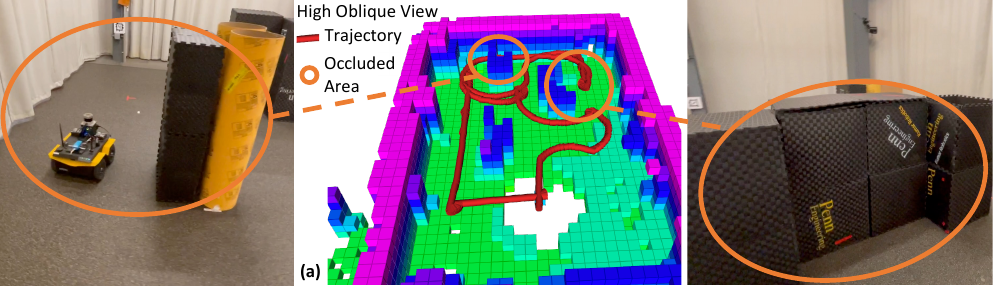}
     \includegraphics[width=\linewidth]{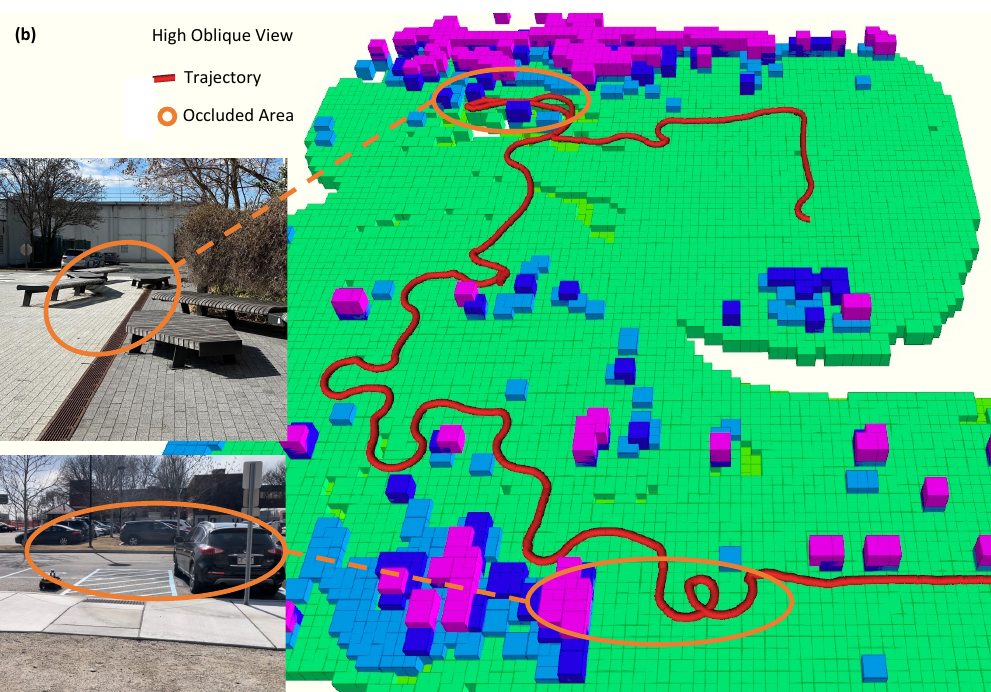}
    \caption{
    \textbf{Reconstructed Occupancy Map and Exploration of Occluded Areas in Real-World Experiment.} 
    The robot navigates to occluded areas, highlighted by orange circles, in indoor (a) and outdoor (b) environments. 
    }
    \label{fig:real_experiment}
\end{figure}

\subsubsection{Setup}
We use a customized Clearpath Jackal platform, details can be found in~\cite{stronger2022Ian}, equipped with an AMD Ryzen 3600 CPU, an Nvidia GTX 1650 GPU and an Ouster OS1-64 LiDAR. We used Faster-LIO~\cite{9718203} for state estimation and move-base~\cite{move-base} for global and local planning. All real-world experiments use the same map representations and definition of information gain as those in~\cref{subsec:dataexp}, except that we use a horizon of $\Delta t = 5$  to reduce the computational load and enable real-time operation. At each timestep, we sampled a 11 $\times$ 11 grid centered around the robot, with a cell size of 2 $\times$ 2 meters, to calculate the information gain at each cell lattice. The path with the maximum specific information gain estimate was obtained using depth first search on the grid, with a maximum length of 10 meters.

\subsubsection{Results}
We carried out experiments in both an indoor environment and an urban outdoor environment. 
One representative result in each environment is shown in~\cref{fig:real_experiment}. 
In the indoor environment in~\cref{fig:real_experiment}(a), the robot actively navigated to the occluded regions behind the barriers. In outdoor environment in~\cref{fig:real_experiment}(b), the robot actively navigated to occluded areas, e.g., the region between benches and the parking lot occluded by the cars.
Areas with high information gain are highlighted. With the proposed approach running online, the robot was able to actively plan paths to maximize the gathered information, resulting in full coverage of areas with high information content.

\section{Conclusion}
Active perception is difficult because the robot has to estimate the actual information that will be gleaned, this gain can only be measured \emph{post hoc}, after observations are realized. This paper demonstrates an online approach to estimate this \emph{post hoc} information gain using the expected information gain. This estimator is critical for instantiating any active perception system. We obtained a mathematical characterization of the improvement in active perception using such an estimator.
We demonstrated our approach for online information gain estimation on a comprehensive set of simulation and real-world experiments. Our approach is capable of building better photometric, geometric and semantic maps using different kinds of robotic perception data in indoor and outdoor environments.

In future work, we wish to deepen our understanding of the design of improvement function classes. Incorporating more complex functions with additional arguments, such as observation history or robot poses, could enhance performance but may also increase the burn-in cost in the regret bound. Additionally, we plan to refine the analysis of regret guarantees. Currently, the regret bound includes a linear term $\lambda T$, as noted in~\cref{rmk:4.1}. By adopting more realistic assumptions about the discrepancy $\Delta r_t$ and exploring more suitable improvement function classes, we may be able to reduce this term to a sub-linear form. Finally, studying the implementation and analysis of randomization for active perception in~\cref{subsec:banditfeedbackregret} could further narrow the gap between theoretical predictions and experimental results.


\section{Acknowledgments}
We gratefully acknowledge the support of NIFA grant 2022-67021-36856, NSF grants 2112665, 2415249, IIS-2145164, CCR-2112665, and the IoT4Ag ERC funded by the National Science Foundation grant EEC-1941529. Siming He acknowledges the support of the Wharton Summer Program for Undergraduate Research.

%% file: lemma_appendix.tex

\section*{Appendix}

\begin{lemma}\label{lemma:4.1}
Follow the regularized leader with a regularization of $\delta_{a}^2/\eta_{a}$ results in the update rule:
\begin{align}\label{eq:updaterule}
    \delta_{a} = \frac{\eta_{a}}{\eta_{a-1}}\delta_{a-1} + \frac{\eta_{a}}{2}d_{a},
\end{align}
where $\eta_{a}$ and $\eta_{a-1}$ are the learning rates for the current and previous updates, respectively.
\end{lemma}
In ``follow the regularized leader'', $\delta_{a} = \arg \min_{\delta} \sum_{i=1}^{t} \left[ \delta(-d_{a}) + \delta^2/\eta_{a} \right]$. We obtain~\cref{eq:updaterule} after finding minimizers $\delta_{a}$ and $\delta_{a-1}$.

\begin{lemma}\label{lemma:4.2}
    \textit{Follow the regularized leader} with regularization $\delta_{a}^2/\eta_{a}$ has the following regret bound.
    \begin{equation}
    \rho_{s,i} \leq \sum^{\alpha^*_{s,i}}_{i=1} \frac{N\eta_{a}}{2} + \frac{\beta^2}{\eta_1}
    \end{equation}
\end{lemma}
We know that $\rho_{s,i} \leq \sum^{\alpha^*_{s,i}}_{i=1}(\delta_{a}-\delta^*)d_{a}$ by convexity. We decompose the bound into two parts: 1) $\sum^{\alpha^*_{s,i}}_{i=1}\delta_{a+1}d_{a}-\delta^*d_{a}\leq\beta^2/\eta_1$ by Lemma 1.4.1 and Lemma 1.4.2 in~\cite{learningingame}; 2) $\sum^{\alpha^*_{s,i}}_{i=1}\delta_{a}d_{a}-\delta_{a+1}d_{a} \leq \sum^{\alpha^*_{s,i}}_{i=1}\vert \delta_{a}-\delta_{a+1}\vert\vert d_{a} \vert\leq \sum^{\alpha^*_{s,i}}_{i=1}N\eta_{a}/2$.

\begin{lemma}\label{lemma:banditregret}
We have the following bound on expected regret in bandit-feedback setting where $\tau$ is the randomization level.
\begin{equation}
    \mathbb{E}(\rho_{j,k})\leq N\beta(N\sqrt{T}+1)/\tau + \tau\beta T
\end{equation}
\end{lemma}
With probability $(1-\tau)$, the full-information algorithm with filled loss gives bound $N^{\Delta t}\beta/\tau(N\sqrt{T}+1)$ by~\cref{thm:4.1}. With probability $\tau$, the regret bound is $\beta T$. Then, we get the expected regret.

\textbf{\textit{Proof Sketch of}}~\cref{thm:4.3}:
Let $\bar{r}_{a}$ be $r_{a,s} + \argmin_{\delta}\sum^{\alpha^*_{s,i}}_{a=1}\left\vert \Delta r_a - \delta \right\vert$ and $\bar{X}_a$ be $\argmax_{X}\bar{r}_{a}(X)$. Let $\hat{r}_a$ be $r_a + \delta_{a}$.
We decompose $r^*_a(X^*_a) - r^*_a(X_a)$ into $[r^*_a(X^*_a) - \bar{r}_a(X^*_a)] + [\bar{r}_a(X^*_a) - \bar{r}_a(\bar{X}_a)] + [\bar{r}_a(\bar{X}_a) - r^*_a(\bar{X}_a)] + [r^*_a(\bar{X}_a) - \bar{r}_a(\bar{X}_a)]  + [\bar{r}_a(\bar{X}_a) - \hat{r}_a(\bar{X}_a)] + [\hat{r}_a(\bar{X}_a) - \hat{r}_a(X_a)]  + [\hat{r}_a(X_a) - \bar{r}_a(X_a)]  + [\bar{r}_a(X_a) - r^*_a(X_a)]$. We have $\mathbb{E}[\sum^T_{t=1}{\hat{r}_t}(X_t) - \bar{r}_t(X_t)]=\mathbb{E}[\sum^{\Delta t}_{s=1}\sum^b_{i=1}\rho_{s,i}]\leq\rho$ and $\sum^{\Delta t - 1}_{s=0}\left(\hat{r}_{t+s}(\bar{X}_{t+s}) - \hat{r}_{t+s}(X_{t+s})\right)=\sum^{\Delta t - 1}_{s=0}[\hat{r}_{t+i}(\bar{X}_{t+s}) - \max_{X}\hat{r}_{t+s}(X)]\leq 0$. Similarly, we can obtain bounds for all components, leading to the inequality in the theorem.

\textbf{\textit{Proof Sketch of}}~\cref{thm:4.4}:
~\cref{thm:4.3} bounds the gap of gain between the optimal paths for specific information gain and for its estimate. This bound is obtained by adding the gap of gain between optimal and approximated paths with estimated specific information gain.